\documentclass{article}
\PassOptionsToPackage{numbers}{natbib}
\usepackage{natbib}
\usepackage{arxiv}
\usepackage{amsmath,amssymb,amsthm}
\usepackage{booktabs}
\usepackage{graphicx}

\theoremstyle{definition}
\newtheorem{definition}{Definition}
\theoremstyle{plain}
\newtheorem{lemma}[definition]{Lemma}
\newtheorem{theorem}[definition]{Theorem}
\newtheorem{corollary}[definition]{Corollary}
\newtheorem{proposition}[definition]{Proposition}
\theoremstyle{definition}
\newtheorem{remark}[definition]{Remark}

\title{World Models Need Not Be Right to Rank Policies\\Right:\\Relative Bounds and Audit Limits}
\renewcommand{\shorttitle}{World Models and Policy Ranking}

\author{
\textbf{Ahanaf Hasan Ariq}\\
\normalfont Ideal School and College
\And
\textbf{Hasaan Ahmad}\\
\normalfont University of the West of England, Bristol
}
\date{}

\begin{document}
\maketitle

\begin{abstract}
Selecting a reinforcement-learning (RL) policy usually requires additional environment interaction, while learned world models offer cheap but potentially biased evaluations. This raises two distinct questions: can shared model errors preserve pairwise rankings despite poor absolute values, and do they make a small number of real audits more informative? We derive an exact pairwise simulation identity and a relative bound separating common-mode Bellman residuals from policy-specific error. The analysis motivates RAVEL (Relative Auditing of Value Errors in Learned models), a selective policy tournament whose eliminations use held-out real outcomes.

We test 45 SAC, TD3, and PPO policies per task in DMC cheetah-run and walker-walk. At rollout horizon 500, model rankings remain strong (Kendall $\tau=0.962$ and $0.800$), and same-run differential-error ratios are $0.674$ and $0.106$. Yet mean paired audit-error correlations are only $0.007$ and $0.090$, and pairing changes standard deviation by $-2.5\%$ and $+1.9\%$. Simultaneous intervals attain $99.98\%$ and $99.75\%$ coverage, but certify only $13.2\%$ and $0\%$ of decisions at eight audits. RAVEL improves low-budget walker-walk regret, whereas real-only successive halving is better at the largest budget, rejecting the precommitted twofold efficiency target. Frozen-model optimization lowers real return without increasing predicted return on average. Thus ranking fidelity, useful audit covariance, and upward model exploitation are distinct properties, separated in part by closed-loop trajectory divergence.
\end{abstract}

\section{Introduction}
Reinforcement learning (RL) has no direct analogue of a supervised validation set. Choosing a checkpoint, seed, algorithm, or hyperparameter setting normally requires additional environment interaction. Off-policy evaluation (OPE) addresses this problem from logged data, but accurate value estimation is difficult under policy-induced distribution shift \citep{thomas2016,jiang2016,fu2021}. Modern world models offer another route: freeze a learned simulator, roll each candidate policy out inside it, and use the resulting return as a cheap validation score.

This use case is no longer hypothetical. WorldEval, WorldGym, and related video-world-model studies report meaningful correlation between simulated and physical policy performance, including preservation of some checkpoint and model rankings \citep{worldeval,worldgym,tseng2025}. A decision-making-centric evaluation proposal consequently places policy ranking near the top of the evidence ladder for useful world models \citep{yu2026}. Yet two issues remain.

First, current evaluations emphasize global statistics such as Pearson correlation or average rank violations. Model selection applies an $\arg\max$: among many candidates it chooses the one with the largest predicted value. Increasing the candidate pool increases the opportunity to select an unusually positive model error. High average correlation can therefore coexist with poor top-1 selection. This is a sequential-decision instance of the optimizer's curse \citep{smith2006}; recent work formalizes both ordinal model exploitation in RL \citep{bhamidipaty2026} and the selection tax for general surrogate-assisted search \citep{ma2026trust}.

Second, standard simulation bounds control each policy's absolute value error separately. They can be unnecessarily pessimistic for ranking. If a model underestimates every policy by the same amount, its absolute error may be large while every pairwise order is exact. Ranking depends on the difference between two model biases, not their sum. The relevant scientific question is thus: \emph{When can an inaccurate world model nevertheless provide a reliable policy ranking, and how much real interaction is needed to discover when it cannot?}

We make four contributions.
\begin{enumerate}
  \item An exact pairwise simulation identity and a relative residual/occupancy bound for ranking and top-1 regret certificates.
  \item RAVEL, which orders comparisons with model rollouts but certifies eliminations only with held-out real outcomes.
  \item A controlled study separating expected-bias cancellation from finite-sample audit covariance.
  \item Evidence that strong rankings need not make matched audits informative at horizon 500.
\end{enumerate}

Equation (2) subtracts two standard simulation identities, and (5) is an elementary triangle-inequality bound; neither step is claimed as a new lemma. Our contribution is the explicit common-residual/occupancy decomposition, its ranking certificates, and evidence that expected-error cancellation need not yield audit savings.

\section{Related Work}
\paragraph{World models and evaluation.}
DreamerV3 and TD-MPC2 use learned dynamics primarily for policy improvement \citep{hafner2025,hansen2024}; compounding error motivates short or conservative model use \citep{janner2019,rigter2022}. WorldEval, WorldGym, and related video-model studies instead evaluate policies and report partial agreement with real rankings \citep{worldeval,worldgym,tseng2025}, but do not give a ranking-specific simulation decomposition or candidate-pool stress test.

\paragraph{Policy selection and confidence.}
OPE estimates values from logged data, including doubly robust estimators \citep{jiang2016}. OPCC benchmarks offline pairwise confidence and model-based baselines on logged data \citep{koul2022}; we do not claim pairwise confidence or model-based comparison as new. We rank fixed policies from frozen world-model rollouts and correct scores with held-out real outcomes. Liu et al. study OPE-model-selection tournaments \citep{liu2025}, while proxy-assisted best-arm methods use correlated predictions as controls \citep{ma2026proxy}. RAVEL is not a generic new tournament: it focuses on a world-model residual/occupancy bound and whether matched errors reduce audit variance. Surrogate methods also require true-data certification \citep{ma2026trust}.

\paragraph{Relative error.}
Bhamidipaty et al. characterize ordinal inversions in imperfect world models \citep{bhamidipaty2026}. Counterfactual quotient models remove action-independent shared future components while preserving action comparisons \citep{chen2026}; our setting instead concerns fixed-policy return ranking under policy-dependent occupancies and real-audit certification. Classical simulation lemmas bound absolute value error \citep{lobel2024}, while cancellation can aid cross-validation even when absolute risk is inaccurate \citep{wager2019}.

\section{Setup}
Let the environment be a discounted Markov decision process $\mathcal{M}=(\mathcal{S},\mathcal{A},P,r,\rho,\gamma)$ with $\gamma\in(0,1)$, initial distribution $\rho$, and bounded rewards. A learned world model $\widehat{\mathcal{M}}=(\mathcal{S},\mathcal{A},\widehat{P},\widehat{r},\rho,\gamma)$ shares the state and action spaces. The analysis also applies in a learned latent space when the policy, reward head, and transition model induce a Markov state.

For a stationary policy $\pi$, write $V^\pi$ and $\widehat{V}^\pi$ for its value functions in $\mathcal{M}$ and $\widehat{\mathcal{M}}$, and
\[
J(\pi)=\mathbb{E}_{s_0\sim\rho}V^\pi(s_0),\qquad
\widehat{J}(\pi)=\mathbb{E}_{s_0\sim\rho}\widehat{V}^\pi(s_0).
\]
The normalized discounted state--action occupancy in the environment is
\[
d^\pi(s,a)=(1-\gamma)\sum_{t=0}^{\infty}\gamma^t
P_{\mathcal{M},\pi}(s_t=s,a_t=a).
\]
It integrates to one. For a bounded function $f$, define
\[
((P-\widehat{P})f)(s,a)=\mathbb{E}_{s'\sim P(\cdot\mid s,a)}f(s')-
\mathbb{E}_{s'\sim\widehat{P}(\cdot\mid s,a)}f(s').
\]
For probability measures $p,q$, let $\operatorname{TV}(p,q)=\sup_A|p(A)-q(A)|$, and let $\operatorname{span}(f)=\sup f-\inf f$.

\begin{definition}[Bellman model residual]
For policy $\pi$, the model residual evaluated on the model value is
\[
\widehat{b}^{\pi}(s,a)=r(s,a)-\widehat{r}(s,a)+\gamma(P-\widehat{P})\widehat{V}^{\pi}(s,a).
\]
The residual is decision-aware: transition error matters only through its effect on the continuation value used by the policy.
\end{definition}

\section{Relative Simulation Theory}
\subsection{An exact pairwise identity}
\begin{lemma}[Simulation identity]
For any stationary policy $\pi$,
\begin{equation}
J(\pi)-\widehat{J}(\pi)=\frac{1}{1-\gamma}\mathbb{E}_{(s,a)\sim d^\pi}\widehat{b}^{\pi}(s,a).
\tag{1}
\end{equation}
\end{lemma}
\begin{proof}
Let $P^\pi$ and $\widehat{P}^\pi$ be the policy-conditioned state kernels. Subtract the two Bellman fixed-point equations and add and subtract $P^\pi\widehat{V}^\pi$:
\[
V^\pi-\widehat{V}^\pi=(I-\gamma P^\pi)^{-1}
\left(r^\pi-\widehat{r}^\pi+\gamma(P^\pi-\widehat{P}^\pi)\widehat{V}^\pi\right).
\]
Let $d_S^\pi$ be the normalized discounted state occupancy. Left-multiplication by $\rho^\top$, together with $(1-\gamma)\rho^\top(I-\gamma P^\pi)^{-1}=d_S^\pi$ and $d^\pi(s,a)=d_S^\pi(s)\pi(a\mid s)$, gives (1).
\end{proof}

For two policies $\pi$ and $\mu$, let $\Delta(\pi,\mu)=J(\pi)-J(\mu)$ and $\widehat{\Delta}(\pi,\mu)=\widehat{J}(\pi)-\widehat{J}(\mu)$. Subtracting (1) yields the exact ranking error
\begin{equation}
\Delta(\pi,\mu)-\widehat{\Delta}(\pi,\mu)=\frac{1}{1-\gamma}
\left(\mathbb{E}_{d^\pi}\widehat{b}^{\pi}-\mathbb{E}_{d^\mu}\widehat{b}^{\mu}\right).
\tag{2}
\end{equation}
Equation (2) is the key object: the world model may be poor in absolute terms yet rank well when the two occupancy-weighted residuals agree.

\subsection{Common-mode cancellation}
\begin{definition}[Relative simulation error]
For any bounded reference residual $c:\mathcal{S}\times\mathcal{A}\to\mathbb{R}$, define
\begin{align}
B_c(\pi,\mu)=\frac{1}{1-\gamma}\Big(&\mathbb{E}_{d^\pi}|\widehat{b}^{\pi}-c|+
\mathbb{E}_{d^\mu}|\widehat{b}^{\mu}-c|\tag{3}\\
&+|\langle d^\pi-d^\mu,c\rangle|\Big).\tag{4}
\end{align}
The tightest member of this family is $B_{\mathrm{rel}}(\pi,\mu)=\inf_c B_c(\pi,\mu)$.
\end{definition}

\begin{theorem}[Relative simulation bound]
For every pair $\pi,\mu$ and every bounded $c$,
\begin{equation}
|\Delta(\pi,\mu)-\widehat{\Delta}(\pi,\mu)|\leq B_c(\pi,\mu).
\tag{5}
\end{equation}
Consequently, if $\widehat{\Delta}(\pi,\mu)>B_c(\pi,\mu)$, then $J(\pi)>J(\mu)$.
\end{theorem}
\begin{proof}
Insert and subtract $c$ in (2):
\[
(1-\gamma)(\Delta-\widehat{\Delta})=
\mathbb{E}_{d^\pi}(\widehat{b}^{\pi}-c)-
\mathbb{E}_{d^\mu}(\widehat{b}^{\mu}-c)+
\langle d^\pi-d^\mu,c\rangle.
\]
Apply the triangle inequality. The order certificate follows from $\Delta\geq\widehat{\Delta}-B_c>0$.
\end{proof}
Choosing $c=0$ recovers a sum of two absolute residual bounds. A nonzero $c$ can be much tighter.

\paragraph{Residual--occupancy certificate.}
Suppose a bounded $c$ satisfies $\|\widehat{b}^{\pi}-c\|_\infty\leq\epsilon_\pi$ and $\|\widehat{b}^{\mu}-c\|_\infty\leq\epsilon_\mu$. Then
\[
|\Delta(\pi,\mu)-\widehat{\Delta}(\pi,\mu)|\leq
\frac{\epsilon_\pi+\epsilon_\mu+\operatorname{span}(c)\operatorname{TV}(d^\pi,d^\mu)}{1-\gamma}.
\]
Indeed, the first two terms of (4) are bounded by $\epsilon_\pi$ and $\epsilon_\mu$. Because $d^\pi-d^\mu$ has zero total mass, subtracting the midpoint of the range of $c$ does not change its inner product with this signed measure. The total-variation inequality gives $|\langle d^\pi-d^\mu,c\rangle|\leq\operatorname{span}(c)\operatorname{TV}(d^\pi,d^\mu)$.

\begin{corollary}[Exact cancellation]
If $\widehat{b}^{\pi}=\widehat{b}^{\mu}=c$ on the two occupancy supports and $\langle d^\pi-d^\mu,c\rangle=0$, then $\Delta(\pi,\mu)=\widehat{\Delta}(\pi,\mu)$ even when both policies have arbitrarily large absolute value error. In particular, any constant common residual cancels because both occupancies have unit mass.
\end{corollary}

\begin{remark}[What the bound predicts]
Ranking should be easiest for nearby checkpoints of one agent: they tend to share representations, model failure modes, and visitation distributions. It should be harder across algorithms or after extensive optimization inside a frozen model, where both the residual and occupancy differences grow. These are direct experimental predictions, not post-hoc explanations.
\end{remark}

\subsection{Selection guarantees from pairwise bounds}
Let $\mathcal{P}=\{\pi_1,\ldots,\pi_K\}$ be a finite candidate pool and let $\widehat{i}=\arg\max_i\widehat{J}(\pi_i)$.
\begin{corollary}[Top-1 regret]
If (5) holds for every pair with bounds $B_{ij}$, then for a true best candidate $i^*$,
\begin{equation}
J(\pi_{i^*})-J(\pi_{\widehat{i}})\leq B_{i^*,\widehat{i}}.
\tag{6}
\end{equation}
\end{corollary}
\begin{proof}
Because $\widehat{i}$ maximizes model value, $\widehat{\Delta}(i^*,\widehat{i})\leq0$. Therefore $\Delta(i^*,\widehat{i})\leq\widehat{\Delta}(i^*,\widehat{i})+B_{i^*,\widehat{i}}\leq B_{i^*,\widehat{i}}$.
\end{proof}
The usual route would bound the regret by the sum of two worst-case absolute errors. Equation (6) shows that only the pairwise differential error between the chosen policy and the true best policy is needed.

\section{RAVEL: Relative Auditing for Policy Selection}
The population quantities above contain the true transition kernel. RAVEL estimates only the pairwise correction needed for finalists, using real outcomes that are held out from world-model and policy training.

\subsection{Paired model-error audits}
Let $X\sim Q$ be a fresh audit condition drawn from a fixed deployment distribution; it includes the initial state, task instance, and controllable exogenous randomness. For policy $i$, let $G_i(X)$ be an environment return and $\widehat{G}_i(X)$ a model return, with $\mathbb{E}_QG_i(X)=J(\pi_i)$ and $\mathbb{E}_Q\widehat{G}_i(X)=\widehat{J}(\pi_i)$. Model randomness can be included in $X$ or averaged conditionally. Define the model error $D_i(x)=G_i(x)-\widehat{G}_i(x)$ and the pairwise differential error
\begin{equation}
D_{ij}(x)=D_i(x)-D_j(x).
\tag{7}
\end{equation}
Using the same draw of $X$ for both policies is not required for unbiasedness---one may instead subtract two independent sample means---but matching can reduce variance when the model has shared condition-specific error:
\[
\operatorname{Var}(D_i-D_j)=\operatorname{Var}(D_i)+\operatorname{Var}(D_j)-2\operatorname{Cov}(D_i,D_j).
\]
Thus a positive common-mode covariance is an asset for ranking even though it is a liability for absolute calibration.

Given $n_{ij}$ fresh i.i.d. matched audit conditions, RAVEL forms the unbiased corrected gap
\[
\widetilde{\Delta}_{ij}=\widehat{J}(\pi_i)-\widehat{J}(\pi_j)+\overline{D}_{ij}.
\]
This is a coefficient-one control-variate estimator: validity does not require an accurate world model, while efficiency requires its condition-specific errors to correlate with real outcomes. Learning a different coefficient requires separate calibration or a method that accounts for coefficient uncertainty \citep{ma2026proxy}. Fix a maximum of $N$ audits per ordered pair. If $D_{ij}\in[-B_{ij},B_{ij}]$, then for every $1\leq n\leq N$ define
\begin{equation}
u_{ij}(n)=B_{ij}\sqrt{\frac{2\log(2K(K-1)N/\delta)}{n}}.
\tag{8}
\end{equation}
Hoeffding's inequality and a union bound give simultaneous coverage over every ordered pair and every $n\leq N$. Therefore pair selection and stopping may depend arbitrarily on earlier audit outcomes. Set $u_{ij}(0)=\infty$. The display treats $\widehat{J}(\pi_i)$ as the exact model expectation. If it is estimated by finitely many imagined rollouts, let $m_{ij}$ be a simultaneous confidence radius for the model gap and use the total radius $w_{ij}=u_{ij}+m_{ij}$; cheap independent model rollouts can make $m_{ij}$ negligible, but it may not be silently omitted.

\begin{proposition}[Selective tournament validity]
Assume all pairwise intervals $[\widetilde{\Delta}_{ij}-w_{ij},\widetilde{\Delta}_{ij}+w_{ij}]$ simultaneously cover the true gaps. After each update, eliminate every $j$ for which some active $i$ has $\widetilde{\Delta}_{ij}-w_{ij}>0$. Then a true best policy is never eliminated. If one policy remains, it is truly best. If the budget ends with an unresolved set $C$, any returned $i\in C$ has the a posteriori guarantee
\[
J(\pi_{i^*})-J(\pi_i)\leq\widetilde{\Delta}_{i^*i}+w_{i^*i}\leq2w_{i^*i},
\]
where $i^*$ is any true best policy.
\end{proposition}
\begin{proof}
On the simultaneous-coverage event, eliminating a true best $i^*$ would require a policy $j$ with a strictly positive lower bound on $J(\pi_j)-J(\pi_{i^*})$, contradicting optimality. Hence $i^*$ remains active, so a singleton survivor must be optimal. If $i\in C$, it has not been eliminated by $i^*$, and therefore $\widetilde{\Delta}_{i^*i}-w_{i^*i}\leq0$. Coverage then gives $J(\pi_{i^*})-J(\pi_i)\leq\widetilde{\Delta}_{i^*i}+w_{i^*i}\leq2w_{i^*i}$. For an unaudited comparison the convention $w_{i^*i}=\infty$ makes the statement valid but uninformative.
\end{proof}
The proposition applies to the candidate set covered by the simultaneous intervals. A model-ranked shortlist is useful for allocating audits, but removing a candidate solely because of its model score voids a full-pool certificate. Thus RAVEL labels a decision as \emph{global} only after every excluded candidate has a covered comparison establishing its inferiority; otherwise the label is \emph{shortlist-conditional}.

\subsection{Procedure}
RAVEL first orders all policies with cheap model rollouts, but never eliminates from model score alone. It audits consequential champion--challenger pairs on fresh conditions, adaptively eliminates only when a real-data lower confidence bound is positive, and returns either a globally certified singleton or an unresolved/shortlist-conditional set. Audit outcomes remain held out from model and policy training. RAVEL-U uses unpaired per-policy intervals; RAVEL-P uses matched pairwise intervals. Their point gaps coincide under equal samples, so pairing can help only through interval width, elimination speed, or audit cost.

\begin{table}[t]
\caption{Precommitted horizon-500 verdicts. H2 is judged by the declared $2\times$ saving rule, not the later dominance exception. Minimum coverage is computed directly from the supplied selection CSV.}
\label{tab:verdicts}
\centering
\begin{tabular}{lccccc}
\toprule
Task & $\tau_b$ & H1 ratio & H2 saving & Min. coverage & H1/H2/H3/H4 \\
\midrule
cheetah-run & 0.962 & 0.674 & 1.000 & 0.9995 & $\checkmark/\times/\times/\checkmark$ \\
walker-walk & 0.800 & 0.106 & 1.333 & 0.9927 & $\checkmark/\times/\checkmark/\checkmark$ \\
\bottomrule
\end{tabular}
\end{table}

\section{Controlled DMC Study}
\subsection{Design and implementation}
We evaluate DMC cheetah-run and walker-walk at horizon 500. For each task, SAC, TD3, and PPO are trained under three seeds for $10^6$ environment steps; five checkpoints per run give 45 policies. A five-member probabilistic dynamics ensemble is fitted to $5\times10^5$ pooled transitions. Each policy receives 128 imagined rollouts and 50 held-out environment episodes. The analysis contains all 990 unordered pairs, and samples 200 candidate pools at each $K\in\{8,16,32\}$ with audit budgets $n\in\{0,2,4,8\}$. Audit outcomes are not used for model or policy training.

Each task requires fitting a five-member ensemble on $5\times10^5$ transitions and up to $45\times128\times500=2.88\times10^6$ imagined model steps, plus 50 held-out episodes per policy. We did not record wall-clock or energy cost; these counts contextualize the two-task scope but do not quantify its compute burden or imply runtime savings.

The supplied run record reports completion without failed stages on PyTorch 2.3.1/CUDA 12.1 using MuJoCo EGL. Matching was checked by resetting every policy from the same stored seed list. A synthetic unit test is not treated as empirical evidence: it injects shared-noise SD 20 and policy-noise SD 2, making correlation 0.99 by construction. For one logged cheetah SAC run, training return rose from 44.5 to 653.0; the cheetah dynamics holdout MSE fell from 0.961 to a best value of 0.125. Analysis-time returns use a different finite-horizon scale, so all comparisons below use only the common analysis definition. The resulting true/model return ranges are $[40.1,115.5]/[40.7,111.8]$ for cheetah-run and $[24.0,171.0]/[25.4,136.5]$ for walker-walk.

\subsection{Ranking accuracy and two kinds of cancellation}
Table~\ref{tab:verdicts} separates expected-bias cancellation from audit-sample covariance. The precommitted H1 threshold was 0.7, and both tasks pass: a world model can therefore be poorly calibrated in absolute value while preserving differences between related policies. Across relationship strata, the differential-error ratios are (0.674, 0.624, 0.608) for cheetah-run and (0.106, 0.165, 0.193) for walker-walk, ordered as same run, different algorithm, and different seed. Same-run cancellation is strongest only on walker-walk; there its inversion rate is 4.4\%, versus 19.1\% for different-seed pairs. The cheetah ordering is inverted, so policy relatedness is not a universal proxy for cancellation.

Matched initial states do not give useful horizon-500 covariance. Mean pairwise audit-error correlations are 0.007 and 0.090. Relative to independent errors, pairing changes standard deviation by $-2.5\%$ on cheetah-run and $+1.9\%$ on walker-walk. On walker-walk, binned error correlation falls from 0.164 to 0.024 as occupancy MMD increases. The interpretation is closed-loop: matching removes initial-condition variation, but different policies rapidly generate different trajectories and policy-specific model errors.

\begin{figure}[t]
\centering
\includegraphics[width=\linewidth]{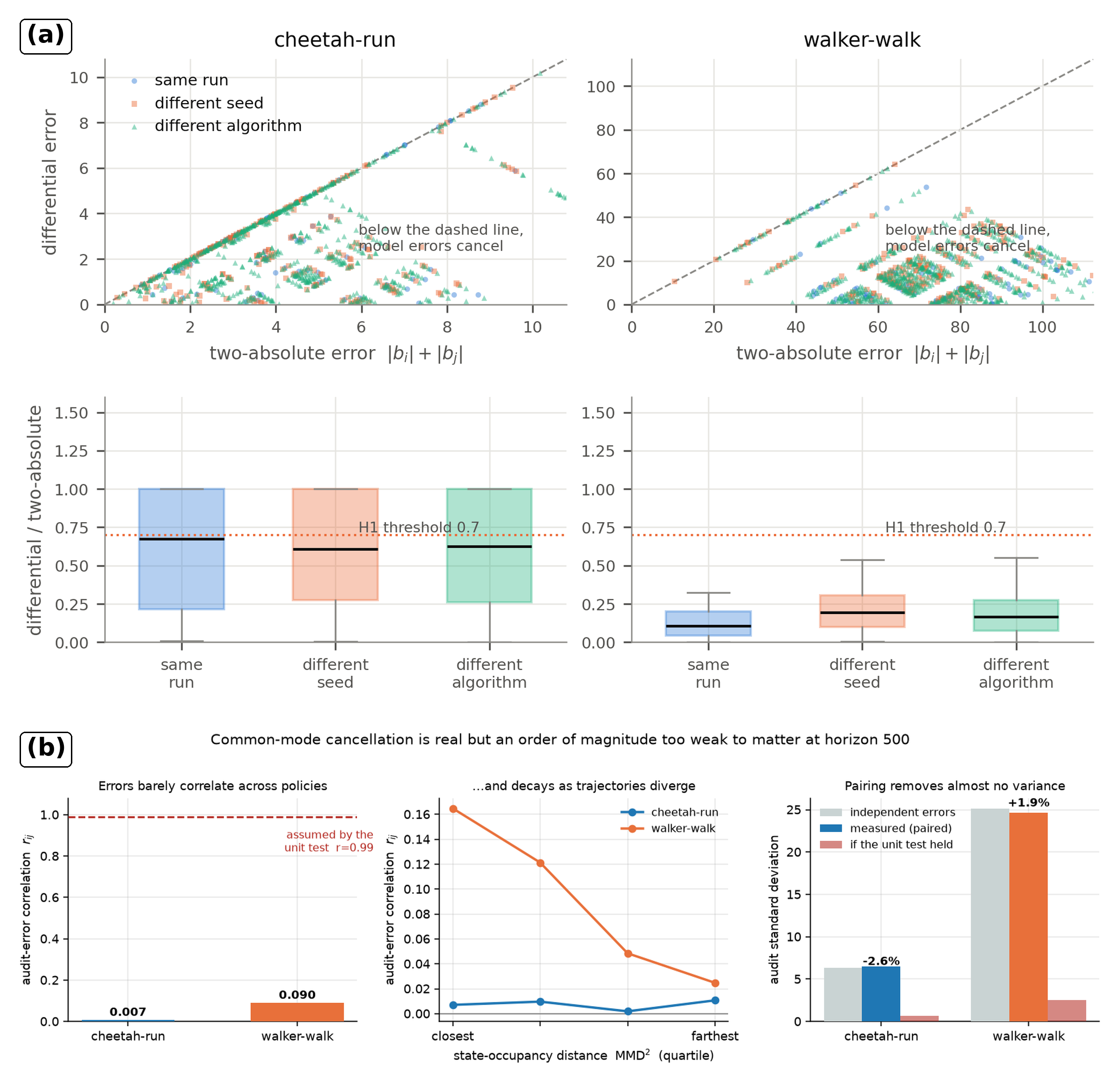}
\caption{Mechanism results. (a) Pairwise differential error is often below the two-absolute-error baseline, although same-run pairs are not uniformly best. (b) Horizon-500 matched audit errors have negligible covariance and yield essentially no variance reduction.}
\label{fig:mechanism}
\end{figure}

\begin{table}[t]
\caption{Normalized top-1 regret from every supplied method/budget condition. Triples are $n=2/4/8$, corresponding to 16/32/64 real episodes. Full precision, abstention, and coverage are in \texttt{results\_selection.csv}.}
\label{tab:regret}
\centering
\begin{tabular}{lcc}
\toprule
Method / audit budget & cheetah-run & walker-walk \\
\midrule
FQE / 0 & .18884 & .13064 \\
Ensemble LCB / 0 & .00261 & .06426 \\
OPCC / 0 & .00261 & .06426 \\
Raw model (also RAVEL at $n=0$) & .00044 & .06878 \\
RAVEL-P / 2,4,8 & .00313 / .00309 / .00008 & .03221 / .01890 / .01833 \\
RAVEL-U / 2,4,8 & .00313 / .00309 / .00008 & .03221 / .01890 / .01833 \\
Real-only SH / 2,4,8 & .01441 / .00480 / .00135 & .03624 / .02784 / .01152 \\
\bottomrule
\end{tabular}
\end{table}
\subsection{Selection, coverage, and abstention}
Raw top-1 regret changes from 0.001073 to 0.000111 as $K$ grows from 8 to 32 on cheetah-run, but rises from 0.047249 to 0.082782 on walker-walk. The aggregate report gives conditional pool-resampling $p$-values of 0.940 and $4.5\times10^{-15}$, respectively. Because the 200 pools overlap and share the same trained policies and evaluator, these values quantify the fixed-suite resampling experiment rather than independent-run generalization. We therefore treat selection pressure as task-specific: cheetah-run is saturated, while walker-walk exhibits the predicted optimizer's-curse pattern.

On walker-walk, RAVEL-P has lower regret than real-only successive halving at $n=2$ (0.03221 versus 0.03624) and $n=4$ (0.01890 versus 0.02784), but successive halving wins at $n=8$ (0.01152 versus 0.01833). The measured episode-saving ratios are 1.00 on cheetah-run and 1.33 on walker-walk. Neither reaches the precommitted twofold threshold, so H2 fails. An implementation clause also marked cheetah-run as passing when successive halving failed to attain a tiny target regret, but that dominance exception was absent from the manuscript criterion; we do not use it.

Empirical coverage exceeds the 93\% floor without post-hoc scaling, so H4 passes literally. Nevertheless, the simultaneous Hoeffding construction is too conservative to be decisive: abstention/certification percentages at $n=2,4,8$ are 100/0, 100/0, and 86.8/13.2 on cheetah-run, and 100/0 at every budget on walker-walk. RAVEL-P and RAVEL-U select the same point winner throughout, as expected from their common corrected means; pairing changes only interval widths, which barely change here.

\begin{figure}[t]
\centering
\includegraphics[width=\linewidth]{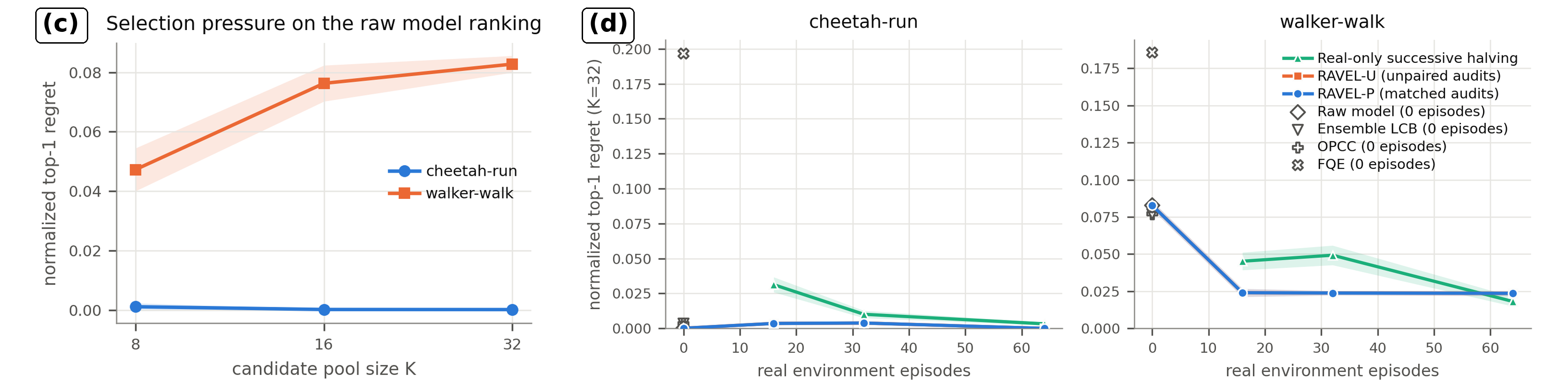}\par
\includegraphics[width=\linewidth]{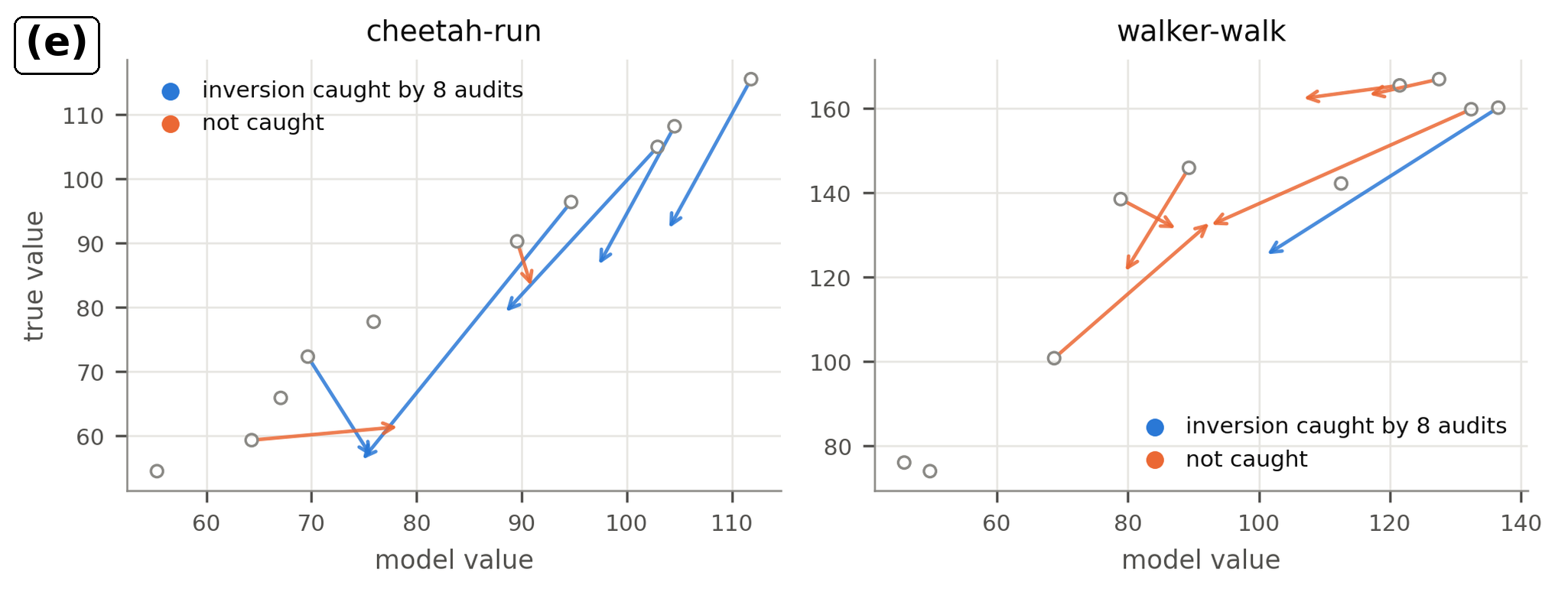}\par
\includegraphics[width=\linewidth]{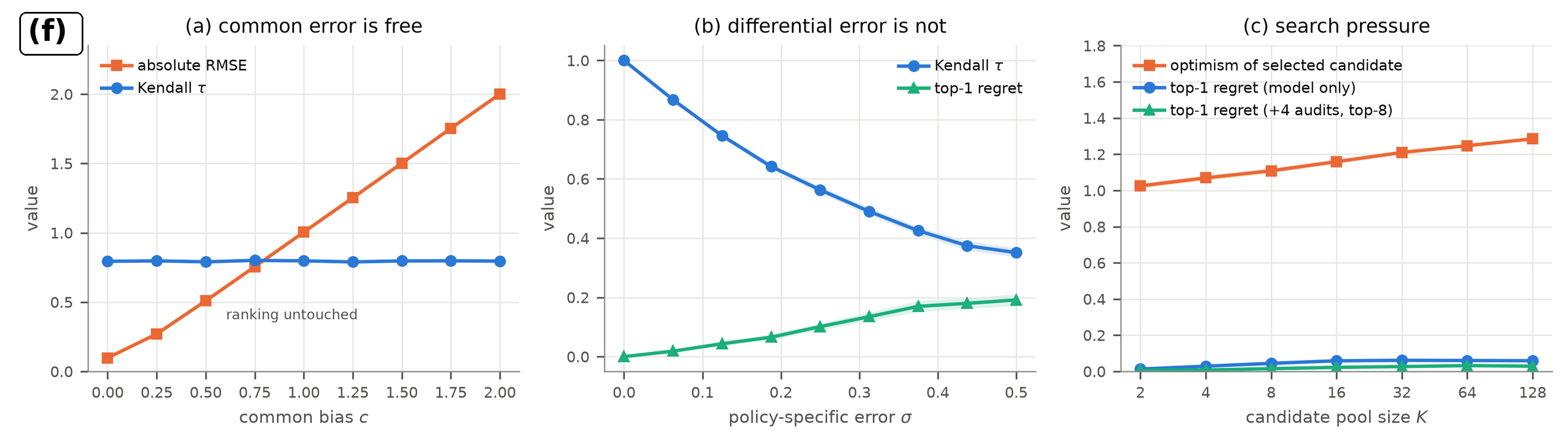}
\caption{Selection and stress tests: (c) raw-model selection pressure; (d) $K=32$ regret; (e) frozen-model updates; and (f) stylized checks of common bias, differential error, and search pressure.}
\label{fig:selection}
\end{figure}

\subsection{Frozen-model stress test}
Ten policies per task receive 500 additional gradient steps against the frozen model. The resulting inversion rates are 0.30 and 0.20, and mean real-return losses are 22.43 and 28.50 for cheetah-run and walker-walk. However, mean model-return changes are also negative, $-4.56$ and $-17.45$. This is not the canonical case in which an optimizer drives predicted reward upward while true reward falls. It is a model-blind collapse: degradation in the environment is substantially larger than, and sometimes invisible to, the model. One cheetah TD3 checkpoint changes from 328.8 to 150.0 in the environment while its model score changes only from 371.6 to 372.9.

Detection at $n=2$ and $n=4$ is 0\% on both tasks. At $n=8$, detection reaches 66.7\% on cheetah-run and remains 0\% on walker-walk. Thus this stress test supplies evidence of consequential ordinal failure but not of systematic upward model exploitation.

\paragraph{Three notions of correctness.}
We distinguish \emph{absolute calibration} ($\widehat{J}_i\approx J_i$), \emph{ordinal utility} (correct differences and selections), and \emph{audit utility} (correlated errors that make small audits informative). The first is unnecessary for the second, and at horizon 500 the second does not imply the third. Evaluation should therefore report differential error and top-policy regret together with audit covariance, coverage, and abstention.

\section{Limitations and Scope}
The identity is exact under the stated Markov/shared-state assumptions, but certificates still require representative audits, bounded errors, and finite-rollout uncertainty; RAVEL cannot remove support mismatch. We study two state-based tasks, one horizon-500 suite per task, overlapping pools from 45 policies, and estimated 50-episode references, so the results establish conditional rather than independent-run generalization. Short horizons, smaller evaluator datasets, latent models, and robots remain untested. The compute counts above do not include wall-clock or energy measurements. The stress test shows degradation rather than systematic upward exploitation \citep{bhamidipaty2026}, and universal proxy savings are impossible \citep{ma2026proxy,ma2026trust}.

\section{Conclusion}
World models can preserve rankings while remaining poor audit control variates. At horizon 500, RAVEL is safe but indecisive; ranking, calibration, and audit utility must be evaluated separately.

\end{document}